\documentclass[5p,times]{elsarticle}

\usepackage{lineno,hyperref}
\usepackage{amssymb,amsmath,array, mathtools}
\usepackage{adjustbox}
\usepackage{makecell}
\usepackage{verbatim}
\usepackage{chngpage}
\usepackage{booktabs}
\usepackage{xspace}

\usepackage{lipsum}
\usepackage{todonotes}
\usepackage{graphicx, subcaption}
\usepackage{bold-extra}

\usepackage{physics}
\usepackage{tikz}
\usepackage{mathdots}
\usepackage{yhmath}
\usepackage{cancel}
\usepackage{color}
\usepackage{siunitx}
\usepackage{array}
\usepackage{multirow}
\usepackage{amssymb}
\usepackage{gensymb}
\usepackage{tabularx}
\usepackage{booktabs}
\usepackage{algorithm}
\usepackage{algorithmicx}
\usepackage{algpseudocode}
\usetikzlibrary{fadings}
\usetikzlibrary{patterns}
\usetikzlibrary{shadows.blur}
\usetikzlibrary{shapes}

\newcommand{\vct}[1]{\ensuremath{\boldsymbol{#1}}}

\newcommand{\argmin}{\operatornamewithlimits{\arg\,\min}}

\newcommand{\myparagraph}[1]{\smallskip \noindent \textbf{#1.}}
\newcommand{\ie}{{i.e.}\xspace}

\newcommand{\etal}{{et al.}\xspace}

\usepackage{pifont}
\newcommand{\yes}{\ding{51}}%
\newcommand{\no}{\ding{55}}%

\modulolinenumbers[5]

\journal{Neurocomputing}

\begin{document}

\begin{frontmatter}

\title{{FADER}: Fast Adversarial Example Rejection}


\author[unipi]{Francesco Crecchi}
\ead{francesco.crecchi@di.unipi.it}
\author[unica]{Marco Melis}
\ead{marco.melis@unica.it}
\author[unica]{Angelo Sotgiu}
\ead{angelo.sotgiu@unica.it}
\author[unipi]{Davide Bacciu}
\ead{bacciu@di.unipi.it}
\author[unica]{Battista Biggio}
\ead{battista.biggio@unica.it}

\address[unipi]{Dipartimento di Informatica,
Universit\`a di Pisa,
Italy.}
\address[unica]{Dip. di Ingegneria Elettrica ed Elettronica,
Universit\`a degli Studi di Cagliari,
Italy.}

\begin{abstract}
Deep neural networks are vulnerable to adversarial examples, i.e., carefully-crafted inputs that mislead classification at test time. Recent defenses have been shown to improve adversarial robustness by detecting anomalous deviations from legitimate training samples at different layer representations - a behavior normally exhibited by adversarial attacks. Despite technical differences, all aforementioned methods share a common backbone structure that we formalize and highlight in this contribution, as it can help in identifying promising research directions and drawbacks of existing methods. The first main contribution of this work is the review of these detection methods in the form of a unifying framework designed to accommodate both existing defenses and newer ones to come. In terms of drawbacks, the overmentioned defenses require comparing input samples against an oversized number of reference prototypes, possibly at different representation layers, dramatically worsening the test-time efficiency. Besides, such defenses are typically based on ensembling classifiers with heuristic methods, rather than optimizing the whole architecture in an end-to-end manner to better perform detection. As a second main contribution of this work, we introduce FADER, a novel technique for speeding up detection-based methods. FADER overcome the issues above by employing RBF networks as detectors: by fixing the number of required prototypes, the runtime complexity of adversarial examples detectors can be controlled.
Our experiments outline up to $73\times$ prototypes reduction compared to analyzed detectors for MNIST dataset and up to $50\times$ for CIFAR10 dataset respectively, without sacrificing classification accuracy on both clean and adversarial data.

\end{abstract}

\end{frontmatter}



\section{Introduction}\label{sec:introduction}

In recent years, Deep Neural Networks (DNNs) achieved state-of-the-art performances in a wide variety of pattern recognition tasks, including (but not limited to) image classification \cite{He2016}, natural language processing \cite{Vaswani2017} and reinforcement learning \cite{Silver2016}. These impressive results, made DNNs appealing for building \textit{smart applications}, i.e. software embedding an \textit{intelligent} component in a form of a DNN to perform pattern recognition, planning, and recommendations.
Today, deep learning is a core component for software spanning from consumer to safety-critical applications like autonomous driving \cite{Bojarski2016}, malware and intrusion detection~\cite{Gibert2020}, homeland security \cite{CarlosRoca2018}, and medical diagnosis \cite{Vaka2020}.

Deep learning (DL) models, as other machine learning algorithms, are designed to work under the so-called \textit{stationarity assumption}: the training data distribution and that of the test samples are assumed to be \textit{the same}. However, distribution drifts can happen \textit{naturally}, e.g., missing data due to sensor failure, or \textit{adversarially}, i.e., an adversary that tampers with data purposely to cause failures during system operation~\cite{huang11,biggio18}.
While DNNs are known to be robust to random noise, it has been shown that the accuracy of DNNs and, in general, of machine-learning algorithms, can dramatically deteriorate in face of gradient-based adversarial attacks~\cite{biggio12-icml,biggio13-ecml}, including adversarial examples, i.e., carefully-perturbed input samples that mislead classification at test time~\cite{szegedy14-iclr,goodfellow15-iclr}. A plethora of methods has been proposed to find adversarial examples \cite{feinman17-arxiv, melis17-vipar, sotgiu20, Papernot2018, meng17-ccs, crecchi19-esann, Lamb2018, Samangouei}. These often transfer across different architectures, enabling black-box attacks even for inaccessible models \cite{papernot16-transf}. The vulnerability of DL models to adversarial samples has the potential to make them the weakest link in the security chain of smart applications. A great number of countermeasures to adversarial examples have been deployed during recent years, still leaving this as an open research problem.

We can categorize defense mechanisms against adversarial examples into two main complementary groups: \textit{robust learning} and \textit{detection} methods~\cite{biggio18}. The former typically employ \textit{adversarial training}~\cite{goodfellow15-iclr}, i.e., retrain the model on adversarial examples to improve classifier robustness against specific attack algorithms. This requires, however, generating attack samples during model training, which may be very computationally demanding for state-of-the-art DNNs. 

Detection approaches, instead, include explicit detection or rejection strategies for adversarial samples, i.e., they provide an additional class for anomalies and potential out-of-distribution attacks. Typically, these defenses are designed to work under the so-called \textit{manifold hypothesis}: in several domains, natural data are assumed to lie in a low-dimensional manifold embedded in a high-dimensional space (e.g., grayscale digits image domain). Remarkably, not every high-dimensional representation belongs to the natural data manifold (e.g., salt and pepper noise).
Assuming adversarial examples to be out-of-manifold data, manifold-based defenses work by identifying adversarial points from their distance to the manifold and, optionally, by ``pulling them back'' onto the data manifold before classification. These defenses are based on a distance-based rejection strategy: as far as a sample moves away from class prototypes, classifier support decreases till zero. If an input sample is not supported by any class, then it is rejected. Remarkable instances of this approach in the literature are found in~\cite{feinman17-arxiv, melis17-vipar, sotgiu20, Papernot2018, Lamb2018, Metzen2017b, crecchi19-esann, meng17-ccs}.
Apart from technical differences, all these rejection-based defenses share a common backbone structure which can be abstracted as a framework. 

The first main contribution of this work is to provide a comprehensive review of such adversarial examples detection methods in the form of a unifying framework. Each proposed detector defense can be obtained by correctly instantiating our framework components.
Subsuming each analyzed detector defense in the framework allowed us to identify common drawbacks, leading us to the second main contribution of this paper.

The vast majority of detector defenses in literature are a form of instance-based classifiers: when a new sample is fed to the classifier, it is compared with a set of prototypes to produce an output prediction. The number of selected training prototypes is thus crucial for the runtime efficiency of the detector, which we found to be not properly tuned in literature solutions. A second issue we found in existing approaches is that, for multilayered defenses, classifiers are optimized to maximize detection separately, i.e. they are not jointly trained to perform rejection.
They are typically based on ensembling classifiers with heuristic methods, rather than optimizing the whole architecture in an end-to-end manner to better perform detection.

To overcome these limitations, in this paper, we propose FADER, a technique for speeding up detection methods. It works by replacing the detector's distance-based classifiers with size-constrained RBF networks, to reduce computational overhead at test time. The proposed solution is capable of enforcing adversarial robustness even in presence of adaptive attacks specifically designed to defeat such defense (see Section \ref{sec:seval}).



In summary, we make the following contributions:
\begin{itemize}
    \item Comprehensive literature review of the detector-based defenses to provide a unified view in the form of a framework, which helps identify current defenses limitations.
    
    \item Overcoming such limitations by proposing FADER, i.e., a technique to obtaining an end-to-end differentiable detector capable of an up to $80\times$ prototypes reduction with respect to analyzed competitors.
    
    \item Novel adaptive attack algorithm designed for the proposed defense method to avoid security by gradient obfuscation.
\end{itemize}

The rest of the paper is structured as follows. In Section \ref{sec:framework} we present our adversarial examples detector framework. 
Section \ref{sect:fader} introduces FADER, our proposed fast detection method.
Section \ref{sec:seval} devises how to perform a reliable security evaluation properly and the adaptive attack specifically designed to evaluate our proposed fast detector. FADER-based solutions are empirically evaluated in Section \ref{sec:experiments}, testing adversarial attack detection in different image recognition tasks. Section \ref{sec:related} discusses complementary methods for addressing adversarial examples than detection strategies. We conclude the paper by discussing the main contributions of this work and its limitations, along with promising future research directions (Section \ref{sec:conclusions}).

\section{Adversarial Examples Detection Framework}\label{sec:framework}

\begin{figure}[t]
    \centering
    \resizebox{0.49\textwidth}{!}{
    \input{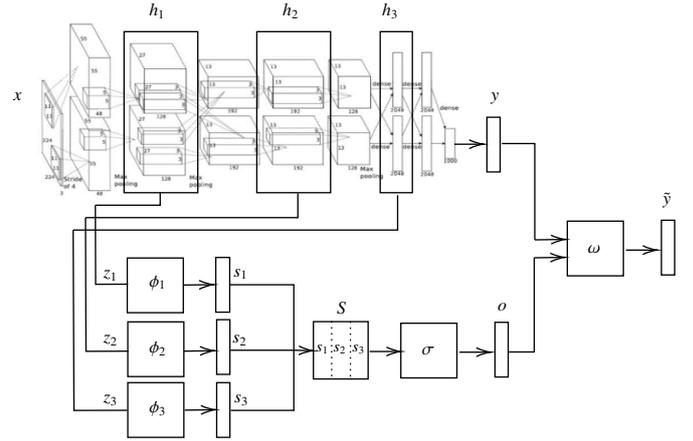}
    }
    \caption{Adversarial examples detection framework architecture. It extends a pre-trained deep network by attaching several layer-detectors $\phi$ whose goal is to determine distribution drifts in the representation of an input $x$ at a given layer. Multiple layer detectors predictions are combined and fed to a combiner classifier $\sigma$ which outputs the final detector prediction. Unprotected network and detector outputs are combined in $\omega$ for final predictions.}
    \label{fig:framework}
    \vspace{-0.3cm}
\end{figure}

In Fig.~\ref{fig:framework} we schematize the proposed detection framework, which assumes an already-trained DNN classifier to be protected against adversarial examples, denoted as a function $f: X \xrightarrow{} Y$ with $X \in \mathbb{R}^d$ and $Y \in \mathbb{R}^c$ for a $d$-dimensional space of input samples (e.g. image pixels) and being $c$ the number of classes. Assuming the network to be composed by $m$ layers, then the prediction function $f$ can be expressed as a series of nested functions $f(h_m(h_{m-1}(\ldots h_1(x; w_1)); w_{m-1}); w_{m})$, where $h_1$ and $h_m$ denote the mapping function learned, respectively, by the input and the output layer, and $w_1$ and $w_m$ are their weight parameters (learned during training).

Detector setup starts by selecting a set of network layers, allowing to inspect internal DNN representations for a given input sample to identify adversarial patterns.

Let $z_i=h_i(h_{i-1}(\ldots(h_1(x; w_1)); w_{i-1}); w_{i})$ be the representation learned by the network of an input sample $x$ at layer $h_i$. A \textit{layer detector} function $\phi_i: Z_i \xrightarrow{} S_i$ is applied to each $z_i$ producing a score vector $s_i$ of size $c$. Intuitively, it represents the probability for $x$ to be a natural sample (i.e. \textit{on-manifold}) for each class, as it appears at $i$-th layer of the network.

Multiple layer detector predictions are combined in the form of a scoring matrix $S$ and processed by the mean of a \textit{multilayer detector} function $\sigma: S_{i}^{k} \xrightarrow{} O$, for $k$ being the number of layers inspected by the detector and $o \in O$ being the detector final prediction score vector. Intuitively, for a given natural sample $x$ of class $t \in Y$, a well-behaving detector is expected to produce reliable, high confidence layer detector predictions for class $t$, namely $S_{i}^{k}[t] \approx 1$, and close to zero scores for other classes. This trend is captured by the multilayer detector which produces a high confidence output score $o_y$ for $x$ to be a natural sample of class $y$. On the contrary, unreliable or low confidence scores $S_{i}^{k}[t]$ produced by layer detectors should drive toward sample rejection, by having the multilayer detector $\sigma$ producing a low confidence output score for each possible output class.

Finally, multilayer detector predictions and original DNN ones have to be combined to produce a final prediction for an input sample $x$. This can be formalized as a function $\omega: Y \times O \xrightarrow{} \Tilde{Y}$, for $y \in Y$ and $o \in O$ being the DNN and the detector predictions respectively, and $\Tilde{y} \in \Tilde{Y}$ being a $c+1$ output vector with an additional rejection class reserved for the detected adversarial examples. Without loss of generality, $\omega$ can be defined as
\begin{equation}
    \omega(y, o, \theta) = [o + \alpha \cdot y;~\theta]    \enspace, 
    \label{eq:default_omega}
\end{equation}
with $\alpha$ being a parameter regulating the contribution of original DNN output for the final predictions $\Tilde{y}$ and $\theta$ being a suitable threshold value tuned on clean training samples.
It is worth to remark that, in all analyzed defenses original DNN predictions are not taken into account for final classification, i.e., $\alpha$ is zero.

For single layer defenses, no combiner is clearly needed. In our framework this corresponds to instantiating $\sigma$ as the identity function. This way, given a single layer detector $\phi_i$ at layer $i$, $S = [s_i]$ and $o = \sigma(S) = s_i$.

In the remainder of the section, we will rephrase existing adversarial examples defenses in terms of the proposed framework. As already mentioned, we are considering only rejection-based defenses against adversarial attacks. To ease the reader, evaluated defenses are schematized in terms of framework components in Table~\ref{tab:defenses_table}.

\subsection{Neural Reject}

Inspired by the concept of \textit{open set} recognition, Melis \etal \cite{melis17-vipar} proposed a method called Neural Reject (NR), which attaches a Support Vector Machine with an RBF kernel (SVM-RBF) on the last hidden layer of a DNN to perform rejection of samples showing an outlying behavior.
In particular, the choice of the RBF kernel implies that the prediction scores provided by the SVM are proportional to the distance of the input sample to the reference prototypes (i.e., the support vectors), thus enabling rejection of samples which fall far away from the training data in the given representation space.
This single-layered defense can be expressed in our framework, instantiating $\phi_{m-1}$ as an SVM-RBF.

\subsection{Kernel Density Estimation}
Feinman \etal \cite{feinman17-arxiv} proposed an adversarial examples detector exploiting a Kernel Density Estimator (KDE) on the embeddings obtained from the last hidden layer of the neural network to identify low confidence input regions. As for NR, such defense can be obtained by instantiating $\phi_{m-1}$ as a KDE.

\subsection{DNN Binary Classifier}
The idea of a layer-wise detector is further developed in \cite{Metzen2017b} providing a single detector subnetwork connected to an arbitrary layer of the DNN which is intended to protect. This subnetwork is trained to perform a binary classification task to distinguish genuine data from samples containing adversarial perturbations. In our framework, $\phi_i$ is the detector subnetwork at a given layer $i$.

\subsection{Dimensionality Reduction}
Multiple-layers inspection has been performed by Crecchi \etal \cite{crecchi19-esann}, who proposed a detection method combining non-linear dimensionality reduction techniques (i.e.~$t$-SNE~\cite{VanDerMaaten2008a}) and density estimation to detect adversarial samples. For a given layer $i$ of the network, the classifier obtained by performing density estimation on top of the embeddings produced by \textit{t}-SNE represents $\phi_i$, whereas the support vector machine combiner is a realization of $\sigma$.

\subsection{Deep Neural Reject}
Sotgiu \etal \cite{sotgiu20}  proposed to apply NR to multiple internal layer representations to form a Deep Neural Rejection (DNR) detector, empirically demonstrating improvements upon the single-layered solution. As for NR, $\phi_i$ is obtained through SVM-RBF classifiers at layer $i$, whereas $\sigma$ is again an SVM-RBF, trained via \textit{stacked generalization} \cite{wolpert92}.

\subsection{Deep k-Nearest Neighbour}
Papernot \& McDaniel proposed a detection method named Deep k-Nearest Neighbour (DkNN) \cite{Papernot2018}, which employs a k-nearest neighbor classifier on the representations of the data learned by each layer of the DNN. When a test input is fed to DkNN, it is compared to its neighboring training points according to the distance that separates them in the representations to estimate the nonconformity, i.e. the lack of support, for a prediction in the training data. If the input sample is not conformed with the training data, it is rejected as an adversarial example. This defense can be obtained by employing kNNs for $\phi_i$ for layer $i$ of the DNN. Statistical hypothesis test for combiner predictions in the realm of \textit{conformal predictions} \cite{Saunders1999, Vovk1999, Shafer2008} can be used as $\sigma$.

\subsection{Generative Models}
As generative models are trained to \textit{approximate} the data generating distribution (which is typically unknown), they are a natural candidate for manifold-based defenses against adversarial examples. Meng \& Cheng proposed MagNet \cite{meng17-ccs} for defending neural network classifiers against adversarial samples leveraging generative models. MagNet works in the input space and employs one or more separate detector networks in the form of a denoising autoencoder (DAE) exploiting the reconstruction error to estimate how far a test sample is from the manifold of normal samples and to \textit{reform} it to a natural sample lying on the data manifold, which is used for classification. 

Fortified Networks \cite{Lamb2018} exploit this very same idea but on the learned hidden representation distribution: DAEs are inserted at crucial points between layers of the original DNN to \textit{clean-up} the adversarial sample lying away from the original data manifold, arguing that this provides stronger protection against adversarial examples than acting in the input space. 

Magnet and Fortified Network defenses can be obtained in our framework by instantiating $\phi_i$ as a DAE, for layer $i$. By having $\sigma$ as the identity function, threshold-based detection ($\omega$) on input sample reconstruction error can be used to identity outliers to the expected input distribution.


However, despite the promising theoretical background, all these methods are still vulnerable \cite{carlini17-aisec, athalye18}.


\begin{table}
\centering
\resizebox{0.45\textwidth}{!}{
\begin{tabular}{ |c|c|c|c| } 
 \hline
 Defense & Adv. Training & $\phi$ & $\sigma$ \\
 \hline\hline
 Feinman \etal \cite{feinman17-arxiv} & \no & KDE &  - \\     
 Melis \etal \cite{melis17-vipar} & \no & SVM-RBF & - \\
 Sotgiu \etal \cite{sotgiu20} & \no & SVM-RBF & SVM-RBF \\
 Papernot \etal \cite{Papernot2018} & \no & k-NN & Statistical Test \\
 Lamb \etal \cite{Lamb2018}{} & \no & DAE & - \\
 \hline
 Metzen \etal \cite{Metzen2017b} & \yes & DNN & - \\
 Crecchi \etal \cite{crecchi19-esann} & \yes & \textit{t-}SNE + KDE & SVM \\
 Meng \& Cheng \cite{Meng2017} & \yes & DAE & - \\
 \hline
\end{tabular}
}
\caption{Detector-based defenses against adversarial examples framed in our proposed detector framework ($-$ for unnecessary components).}
\vspace{-0.5cm}
\label{tab:defenses_table}
\end{table}

\section{Fast Adversarial Example Rejection} \label{sect:fader}

In this section, we present our proposal for speeding-up existing detection methods for adversarial examples by controlling the number of reference prototypes they make use of. Previous instance-based detectors~\cite{melis17-vipar, sotgiu20, feinman17-arxiv, Papernot2018, crecchi19-esann}, in fact, do not allow one to specify the number of prototypes (e.g. \textit{support vectors} for SVM-based ones) used for identifying adversarial examples. Selecting a large number of reference prototypes, possibly at different representation layers, dramatically increases classification time, as the input sample has to be compared with each prototype at each selected representation layer to compute the corresponding prediction. Thus, controlling the number of prototypes employed by detectors is crucial for runtime efficiency. With FADER, we propose to replace existing classifiers in such detectors with size-constrained RBF networks designed for an optimal accuracy vs. speed tradeoff.

RBF networks are \emph{shallow} artificial neural networks that use radial basis functions (RBF) as activation functions. The output of the network is a linear combination of radial basis functions of the inputs and neuron parameters. Despite their architectural simplicity, they have been shown to possess structural resistance to adversarial attacks \cite{goodfellow15-iclr, DeAlfaro2018, Habib2019, Chenou2019a}, thanks to their localized nature, thus they are a natural candidate for building fast and secure detectors. The use of RBF activation functions enforces the classifier to assume a desirable \emph{compact abating probability} property for open set recognition \cite{bendale16-cvpr, scheirer11-pami}. Being $s_1, \ldots, s_c$ the output scores produced by the classifier for an input sample $x$, such property ensures that for each given class, scores decrease while $x$ moves away from input regions densely populated by training samples of that class. This property allows us to easily implement a \textit{distance-based rejection} mechanism as the one required in our case to detect adversarial examples, as in \cite{melis15-iciap, bendale16-cvpr, sotgiu20}. 

\subsection{FADER}
The central idea of FADER is to speedup instance-based adversarial examples detectors by controlling the number of prototypes used for comparison while maintaining at least comparable performances with original solutions. To this end, we replace detector classifiers with RBF network-based ones, allowing for the joint optimization of prototypes and network parameters. Suppose to take NR as a reference detector we intend to speedup, given $z_{m-1}$ as the input representation obtained at the logits layer of the DNN, NR decision function can be formulated as follows:
\begin{equation}
	f(z_{m-1}) = sgn \left( \sum_{i=1}^n y_i\alpha_i~\text{exp}\left( - \frac{|| z_{m-1} - z_i||^2}{\gamma}\right) + b \right)
\end{equation}
that is, classifier output is computed by comparing $z_{m-1}$ with each support vector $z_i$ (identified by $\alpha_i>0$). Remarkably, the number of support vectors is automatically determined by the SVM training procedure, in a data-driven way, and reference prototypes $z_i$ are drawn from training samples, meaning they are not optimized by the training procedure. Kernel bandwidth $\gamma$ is typically determined using cross-validation and not tuned for each reference prototype separately. As a result, over-specified solutions in terms of deployed prototypes are typically obtained (as in Section \ref{sec:exp-results}).

\begin{figure*}[t]
\centering
\subcaptionbox{}{\includegraphics[width=.24\textwidth,trim=0 0 8 0, clip]{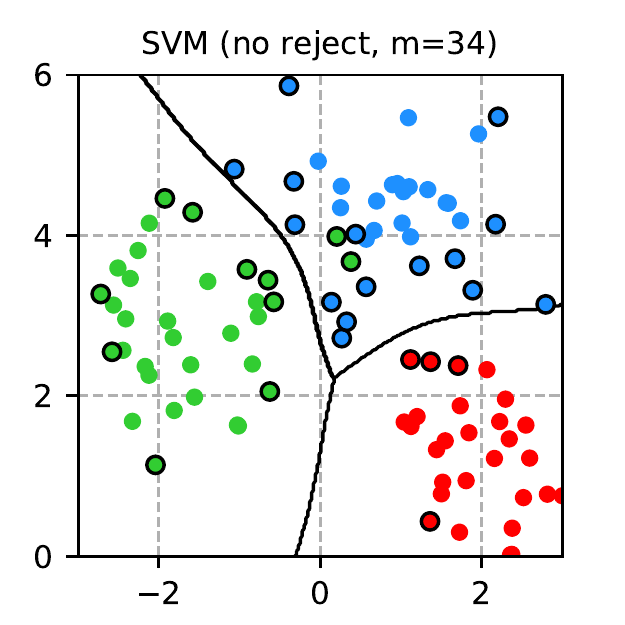}}
\subcaptionbox{}{\includegraphics[width=.24\textwidth,trim=0 0 8 0, clip]{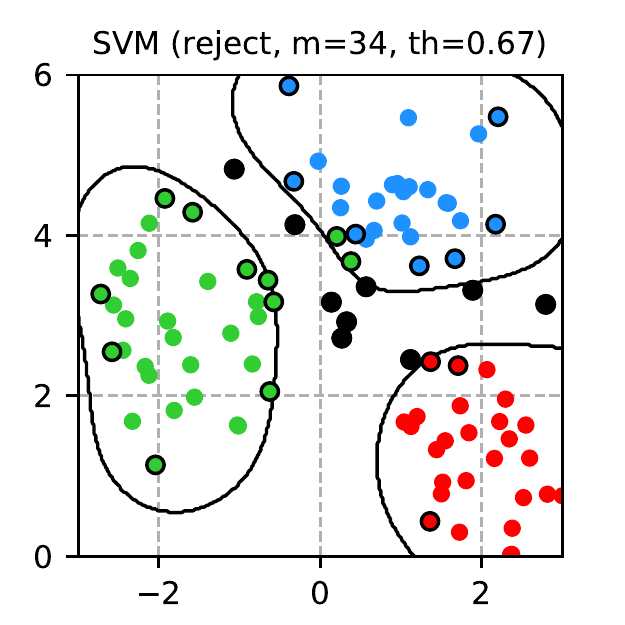}}
\subcaptionbox{}{\includegraphics[width=.24\textwidth,trim=0 0 8 0, clip]{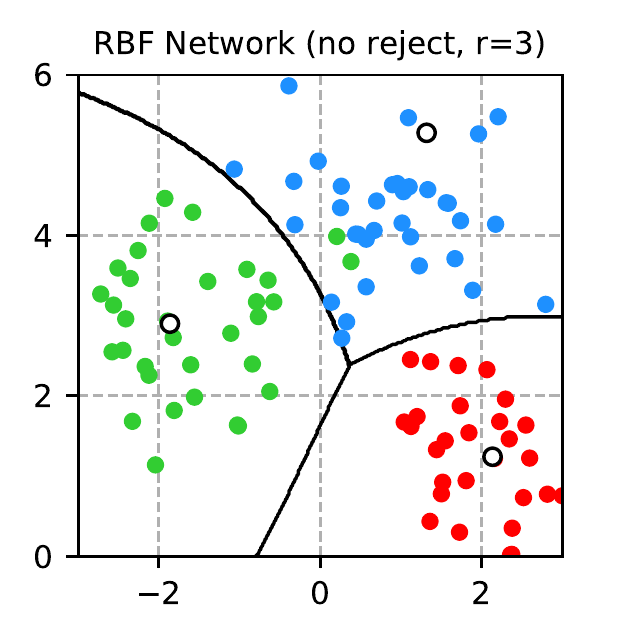}}
\subcaptionbox{}{\includegraphics[width=.24\textwidth,trim=0 0 8 0, clip]{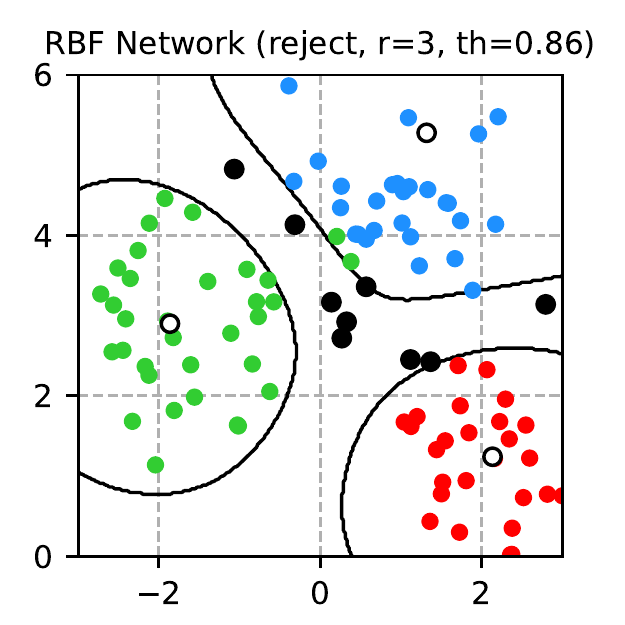}}
\vspace{-.5em}
\caption{Comparison of classifiers decision regions on a two-dimensional classification example with three classes (green, blue, and red points), using multiclass SVMs with RBF kernels (SVM) and RBF Networks. a) SVM without reject option, the solution found exploits $m=34$ support vectors (circled in black). b) SVM with reject option using a threshold $th=0.67$ to obtain 10\% FPR, rejected samples are highlighted with black dots. c) RBF Network without rejecting option, the solution found properly separates all classes using only $r=3$ bases (black circles). d) RBF Network with reject option using a threshold $th=0.86$ to obtain 10\% FPR, rejected samples are highlighted with black dots. Notably, $r=3$ is the minimum number of bases to ensure each class is correctly enclosed.}
\label{fig:toy-example}
\vspace{-1em}
\end{figure*}

Replacing the NR classifier with an RBF network can be beneficial in reducing the number of prototypes employed, while maintaining the desired detector behaviour, i.e., nearly the \textit{same} decision regions of unoptimized solutions (see Fig.~\ref{fig:toy-example}). The new detector decision function can be formulated as follows:
\begin{equation}
	f(z_{m-1}) = sgn \left( \sum_{i=1}^n w_i ~\text{exp}\left( - \frac{|| z_{m-1} - z_i||^2}{\gamma_i} \right) + b \right)
\end{equation}
Although the two definitions really look alike, they substantially differ in practical terms. Reference prototypes $z_i$ can now be tuned during RBF network training procedure, as well as kernel bandwidths $\gamma_i$ in order better fit training data. This improved flexibility allows to control the number of reference prototypes, e.g., to speed-up kernel computation. Moreover, optimizing prototypes ($z_i$), kernel bandwidths ($\gamma_i$) and network parameters ($w_i, b$) all-together, allows for accounting prototypes reduction, while maintaining comparable performances with respect to overspecified solutions, as demonstrated in our experiments (see Section \ref{sec:experiments}).

In terms of the proposed adversarial detector framework in Section \ref{sec:framework}, FADER can be represented as follows: the layer detector function $\phi$ can be instantiated as an RBF network. In the case of multilayered detectors, the combiner $\sigma$ can be instantiated again as an RBF network. Default rejection-based strategy (see Eq. \ref{eq:default_omega}) is employed to mark adversarial examples.

\section{Adversarial Robustness Evaluation}\label{sec:seval}
A correct evaluation of proposed detection methods against adversarial examples is essential \cite{biggio18, athalye18}, and it is not sufficient to evaluate such defenses against previous defense-unaware attacks that are likely to fail (see, e.g. \cite{lu17-iccv, papernot16-distill, meng17-ccs}), leading to overly optimistic results in term of classifier robustness. To perform a fair defense evaluation, attacks should take into account the defense mechanism. Under this condition, many defenses were shown not to be as effective as claimed \cite{carlini17-aisec, carlini17-sp, athalye18}.
For instance, many defenses take advantage of \textit{gradient masking}, i.e. they learn functions which are harder to optimize for gradient-based attacks; however, they can be easily bypassed by constructing smoother, differentiable approximations of their functions, e.g., learning surrogate models \cite{biggio18, biggio13-ecml, russu16-aisec, melis18-eusipco} or replacing network layers which obfuscate gradients with smoother mappings \cite{carlini17-aisec, carlini17-sp, athalye18}. 

\begin{figure}
\centering
\includegraphics[width=.195\textwidth,trim=0 0 1.2cm 0, clip]{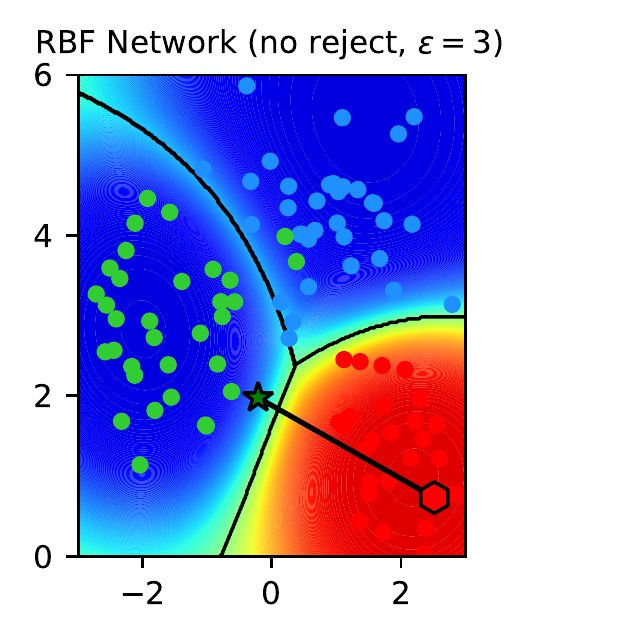}
\includegraphics[width=.241\textwidth]{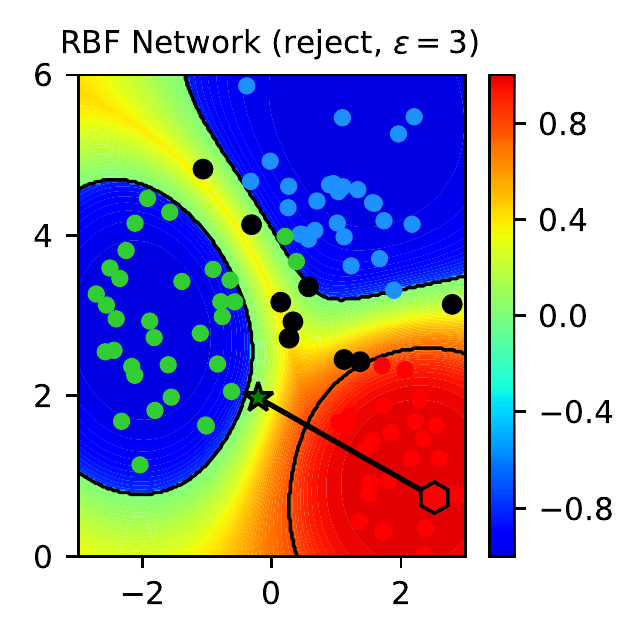}
\vspace{-0.25em}
\caption{Conceptual representation on a 3-class bi-dimensional classification problem of the evasion attack against an RBF Network without (left) and with reject-based defense (right). The initial sample $x_0$ (red hexagon) has been modified to obtain the adversarial sample $x^\star$ (green star) which is wrongly recognized as an observation from the green class by the standard classifier, while it is correctly rejected when employing our defense. The values of the attack objective $\Omega(\vct x)$ are shown in the background for both cases.}\label{fig:evasion2d}
\vspace{-1em}
\end{figure}

When a defense exploits rejection, a defense-unaware attack may craft adversarial examples belonging to rejection regions, making it very difficult to evade such defense (Fig. \ref{fig:evasion2d}). To perform a fair robustness evaluation of the proposed defense method an adaptive defense-aware attack is required.

In this work we adopt the security evaluation procedure proposed in \cite{biggio18, athalye18, biggio14-tkde}, constructing \textit{security evaluation curves} which show how the accuracy of attacked systems degrades under attacks crafted with increasing strength, \ie the amount of perturbation. The more robust a defense method is, the more gracefully the curve decreases. It is remarkably important that accuracy curves should reach zero under a sufficiently large perturbation, for all evaluated defense - in an extreme ideal case, with an unbounded perturbation, the attacker can replace the source sample with a sample of another class. If this does not happen, then it may be that the attack algorithm is not able to perform correctly the optimization, and this in turn means that we are probably overestimating the defense robustness.

We formulate here an adaptive white-box attack suitable for all defenses considered in this work, and that takes into account rejection. Starting from a sample $\vct x$, the attacker can compute a maximum-allowed $\varepsilon$-sized adversarial perturbation obtaining the adversarial example $\vct x^\star$, by solving the following constrained optimization problem:
\begin{equation}
     \vct x^\star \in  \argmin_{\vct x^\prime : \| \vct x- \vct x^\prime\| \leq \varepsilon} \Omega(\vct x)  = s_y( \vct x^\prime) - \max_{j \not \in \{0, y\} } s_j(\vct{x^\prime})
\end{equation}
being $\vct x^\prime$ the modified input sample, $\| \vct x- \vct x^\prime\| \leq \varepsilon$ is an $\ell_p$-norm constraint (typical norms used for crafting adversarial examples are $\ell_1$, $\ell_2$ and $\ell_\infty$, for which efficient projection algorithms exist~\cite{duchi08}), $y \in Y$ is the true class, and $0$ is the rejection class.

Intuitively, to perform an untargeted (error-generic) evasion the output of the true class must be minimized, and the output of one competing class (excluding the reject class) must be maximized. The resulting objective function is negative in case of successful evasion, and its absolute value increases with the increasing classification confidence on the competing class. The algorithm does not simply search for a minimum-distance adversarial example, but it maximizes the confidence of the attack.
Although in this work we focus only on untargeted attacks, the proposed formulation can be easily extended to account for targeted (error-specific) evasion, as in~\cite{melis17-vipar}.

To solve the optimization problem above, we use a projected gradient descent (PGD) algorithm with variable step size, as given in Algorithm~\ref{alg:evasion}. The initial step size $\eta_{0}$ is doubled for ten times, computing the objective function at each step and choosing the step size which minimizes the function. The selected step size is then used to update the point $\vct x^\prime$. Choosing a different $\eta$ at each step gives two main advantages:
\begin{itemize}
	\item Speeding-up the optimization: using larger step sizes (when possible) allows us to reach the convergence with a reduced number of iteration steps.
	\item Escaping local minima which may hinder the optimization process, using the larger step sizes.
\end{itemize}

While attacking DNR, we found that the optimization often got stuck in local minima inside reject decision regions, where the objective function gradient reaches very small values close to zero. 
The magnitude of these gradients strongly depends on the value of the $\gamma$ parameter of SVM-RBF classifiers used by DNR, which is a negative exponent used in the kernel computation that controls the shape of decision regions around training samples. Larger values of $\gamma$ produce more complex decision regions and smaller gradient magnitude, smaller values of $gamma$ conversely produce smoother decision regions. 

To overcome this limitation we introduce \emph{gamma smoothing}, i.e., we perform the attack on a surrogate DNR classifier that computes the gradient using a smaller $\gamma$ which is easier to attack, as discussed in ~\cite{demontis19-usenix}.

\begin{algorithm}[t]
	\caption{PGD-based Maximum-confidence Adversarial Examples with exponential line search}
	\label{alg:evasion}
	\begin{algorithmic}[1]
		\Require $\vct x_{0}$: the input sample;  $\eta_{0}$: the initial step size; $\Pi$: a projection operator on the $\ell_p$-norm constraint $\| \vct x_0- \vct x^\prime\| \leq \varepsilon$; $t > 0$: a small positive number to ensure convergence. 
		\Ensure $\vct x^{\prime}$: the adversarial example.
		\State $\vct x^{\prime} \gets \vct x_{0}$
		\Repeat
		\State $\vct x \gets \vct x^{\prime}$
		\State $\vct x''_{0} \gets \Pi \left ( \vct x -  \eta_{0} \nabla \Omega(\vct x) \right )$
		\For{$k = 1$ \textbf{to} $k < 10$}
			\State $\eta_{k} \gets 2^k \eta_{0}$
			\State $\vct x''_{k} \gets \Pi \left ( \vct x -  \eta_{k} \nabla \Omega(\vct x) \right )$
			\If{$\Omega (\vct x''_{k}) < \Omega (\vct x''_{k - 1})$}
				\State $\eta^\prime \gets \eta_{k}$
			\EndIf
		\EndFor
		\State $ \vct x^\prime \gets  \Pi \left ( \vct x -  \eta^\prime \nabla \Omega(\vct x) \right ) $
		\Until{$ | \Omega (\vct x^\prime) - \Omega (\vct x) | \le t $}
		\State \Return $\vct x^\prime$
	\end{algorithmic}
\end{algorithm}

\section{Experimental analysis}\label{sec:experiments}
In this section, we empirically evaluate the security of the proposed FADER defense mechanism (i) against defense-aware adversarial examples, and (ii) in a black-box setting where the attacker is essentially unaware of the defense mechanism used to protect the DNN classifier. After detailing our experimental setup (Sec. \ref{sec:exp-setup}), we report the classifier's performance under attack by comparing it with NR and DNR detectors (Sec. \ref{sec:exp-results}).

We use \texttt{secml}~\cite{melis19-arxiv} as a Python framework to implement the classification systems and the attack algorithms, planning to extend it soon to include an implementation of FADER-based detectors presented hereafter.

\subsection{Experimental setup}
\label{sec:exp-setup}
We discuss here the datasets we use to evaluate our defense method and the classifiers we compare the performance with.

\myparagraph{Datasets}
Our analysis is performed on MNIST handwritten digits and CIFAR10 datasets. MNIST consists of 60,000 training and 10,000 test gray-scale samples of shape 28x28. CIFAR10 consists of 50,000 training and 10,000 test RBG samples of shape 32x32. All images are normalized in $[0, 1]$ by dividing the input pixel values by 255.

\myparagraph{Classifiers}
We compare the performance of an undefended DNN (i.e. not implementing any rejection mechanism), which represents our baseline, with NR and DNR defense methods and their \textit{fast} variants, i.e. FADER technique is applied.

To implement the undefended DNNs for the MNIST dataset, we use the same architecture suggested by Carlini \etal \cite{carlini2017adversarial}. For CIFAR10, instead, we consider a lightweight network that, despite its size, allows obtaining high performances. The two architectures under consideration are shown in Tables \ref{tab:nn-minst} and \ref{tab:nn-cifar}.

As for the detectors, we consider the single-layer rejection mechanism on top of the pre-softmax activation layer, in the form of an SVM-RBF for NR \cite{melis17-vipar} and the DNR defense approach \cite{sotgiu20} employing SVMs with RBF kernel as both layer detectors and the top combiner. For both NR and DNR we provide \emph{fast} variants, employing size-controlled RBF networks, denoted as NR-RBF and DNR-RBF, respectively (more in Section \ref{sec:param_setup}).

For experimental sake, we limit the number of inspecting layers for deep detectors to three, by considering the last three layers for the network trained on MNIST, and the last layer plus the last batch norm layer and the second to the last max-pooling layer for the one trained on CIFAR10 (choice aimed at obtaining a reasonable amount of features). To ease the reader, we marked the selected layers in bold in Table \ref{tab:nn-minst} and \ref{tab:nn-cifar}.

\begin{table}[t]
    \centering
	\begin{tabular}{llc}
		\toprule
		Id & Layer Type & Dimension\\
		\midrule
		\texttt{relu1} & Conv. + ReLU & 64 filters (5x5) \\
		\textbf{\texttt{relu2}} & \textbf{Conv. + ReLU} & \textbf{64 filters (3x3)} \\
		\textbf{\texttt{relu3}} & \textbf{Conv. + ReLU} & \textbf{64 filters (3x3)} \\
		\textbf{\texttt{relu4}} & \textbf{Fully Connected + ReLU} & \textbf{32 units} \\
		\texttt{dropout} & Dropout ($p=0.5$) \\
		\texttt{softmax} & Softmax  & 10 units \\
		\bottomrule
	\end{tabular}
    \caption{Model architecture of the MNIST Neural Network                \cite{carlini2017adversarial}. The layers used by DNR and FADER detectors are highlighted in bold.}
    \label{tab:nn-minst}
    \vspace{-0.5cm}
\end{table}

\begin{table}[t]
    \centering
    \resizebox{\columnwidth}{!}{%
	\begin{tabular}{llc}
		\toprule
		Id & Layer Type & Dimension \\
		\midrule
		\texttt{relu1} & Conv. + Batch Norm. + ReLU & 64 filters (3x3) \\
		\texttt{relu2} & Conv. + Batch Norm. + ReLU & 64 filters (3x3) \\
		\texttt{drop1} & Max Pooling + Dropout ($p=0.1$) & 2x2 \\
		\texttt{relu3} & Conv. + Batch Norm. + ReLU & 128 filters (3x3) \\
		\texttt{relu4} & Conv. + Batch Norm. + ReLU & 128 filters (3x3) \\
		\texttt{drop2} & Max Pooling + Dropout ($p=0.2$) & 2x2 \\
		\texttt{relu5} & Conv. + Batch Norm. + ReLU & 256 filters (3x3) \\
		\texttt{relu6} & Conv. + Batch Norm. + ReLU & 256 filters (3x3) \\
		\textbf{\texttt{drop3}} & \textbf{Max Pooling + Dropout ($\mathbf{p=0.3}$)} & 2x2 \\
		\textbf{\texttt{relu7}} & \textbf{Conv. + Batch Norm. + ReLU} & 512 filters (3x3) \\
	    \texttt{drop4} & Max Pooling + Dropout ($p=0.4$) & 2x2 \\
		\textbf{\texttt{linear}} & \textbf{Fully Connected} & 512 units \\
		\texttt{softmax} & Softmax & 10 units \\
		\bottomrule
	\end{tabular}
	}
    \caption{Model architecture of the CIFAR10 Neural Network. The layers used by DNR and FADER detectors are highlighted in bold.}
    \label{tab:nn-cifar}
    \vspace{-0.3cm}
\end{table}

\myparagraph{Training-test splits}
We assume the DNNs used in our experiments to be pre-trained on separate training sets of 30,000 MNIST digits and 40,000 CIFAR10 samples. The rest of the data is then used for training the NR and DNR classifiers in both the SVM-based and RBF neurons-based configurations. We average the security evaluation results on five different runs. In each run, we consider 10,000 training samples and 1000 test sample, randomly drawn from the corresponding datasets. As previously done in \cite{sotgiu20}, to avoid overfitting, the DNR combiner is trained by concatenating the outputs of the base SVMs detectors computed on separate validation sets, extracted from the training set using a 3-fold cross-validation procedure. This procedure is known as \emph{stacked generalization}~\cite{wolpert92}.

\myparagraph{Parameter setting}\label{sec:param_setup}
Table \ref{tab:nn-params} reports the hyperparameters used for DNN pre-training.
DNR detectors' best configuration is looked for in $C \in \{10^{-2}, \ldots, 10^{2}\}$ and $\gamma \in \{10^{-4}, \ldots, 10^{2}\}$ performing a 3-fold cross-validation procedure to maximize accuracy on unperturbed training data. The optimal configurations we found for MNIST and CIFAR10 datasets are reported in Table~\ref{tab:dnr-bestconfig}. As DNR layer classifiers and combiners are not independently optimized during training, NR best configuration can be obtained by lookup Table~\ref{tab:dnr-bestconfig} for the layer of interest.

FADER-based solutions architectures are designed to maximize prototype reduction rate, while achieving comparable performances on clean test samples (see Tables \ref{tab:mnist-proto} and \ref{tab:cifar10-proto}). In terms of training, RBF neurons based solutions are optimized using \texttt{pytorch}\footnote{Website: https://pytorch.org/} Adam optimizer with default settings for 250 epochs. 
Rejection threshold $\theta$ is set, in all the considered cases, to reject $10\%$ of the samples when no attack is performed (at $\varepsilon = 0$).

\begin{table}[!ht]
    \centering
	\begin{tabular}{lll}
		\toprule
		Parameter & MNIST & CIFAR10 \\
		\midrule
		Learning Rate & 0.1 & 0.01 \\
		Momentum & 0.9 & 0.9 \\
		Dropout & 0.5 & (see Table~\ref{tab:nn-cifar}) \\
		Batch Size & 128 & 100 \\
		Epochs & 50 & 75 \\
		\bottomrule
	\end{tabular}
    \caption{Parameters used to train the MNIST and CIFAR10 DNNs.}
    \label{tab:nn-params}
	\vspace{-1em}
\end{table}
    

\begin{table}[ht]
    \centering
    \begin{tabular}{ cc }   
        MNIST & CIFAR10 \\  
        \begin{tabular}{lll}
        		\toprule
        		Layer & C & $\gamma$ \\
        		\midrule
        		\texttt{relu2} & 10 & 1e-3 \\
        		\texttt{relu3} & 10 & 1e-2 \\
        		\texttt{relu4} & 1.0 & 1e-2\\
        		\texttt{combiner} & 1e-1 & 1.0 \\
        		\bottomrule
    	\end{tabular} &  
        \begin{tabular}{lll}
        		\toprule
        		Layer & C & $\gamma$ \\
        		\midrule
        		\texttt{drop3} & 10 & 1e-3 \\
        		\texttt{relu7} & 1.0 & 1e-3 \\
        		\texttt{linear} & 1.0 & 1e-2\\
        		\texttt{combiner} & 1e-4 & 1.0 \\
        		\bottomrule
    	\end{tabular} \\
    \end{tabular}
\caption{DNR configurations for MNIST (left) and CIFAR10 (right) datasets.}
\label{tab:dnr-bestconfig}
\end{table}



\myparagraph{Security Evaluation}
We compare the aforementioned undefended neural networks and the rejection-based architectures in terms of their security evaluation curves \cite{biggio18}, reporting classification accuracy against an increasing $\ell_2$-norm perturbation size, used to perturb all the test samples. In particular, under attack (i.e., for $\varepsilon > 0$), all the tested samples are adversarial examples and are considered correctly classified if either assigned to the rejection class or to their true class. We consider a significant interval of $\varepsilon \in [0, 5]$ and $\varepsilon \in [0, 2]$ for the attacks against MNIST and CIFAR10 datasets, respectively. The rejection rates, computed by dividing the number of rejected samples by the number of test samples, are also reported for all defense-aware classifiers (in absence of defense, the rejection rate will be zero). It is worth noting that, under this setting, the fraction of adversarial examples at each $\varepsilon > 0$ which are correctly assigned to their true class is given by the difference between the accuracy and the rejection rate.

\subsection{Experimental results}
\label{sec:exp-results}

\begin{figure}[t!]
    \centering
    \fbox{\includegraphics[width=.45\textwidth, trim=40 10 50 10, clip]{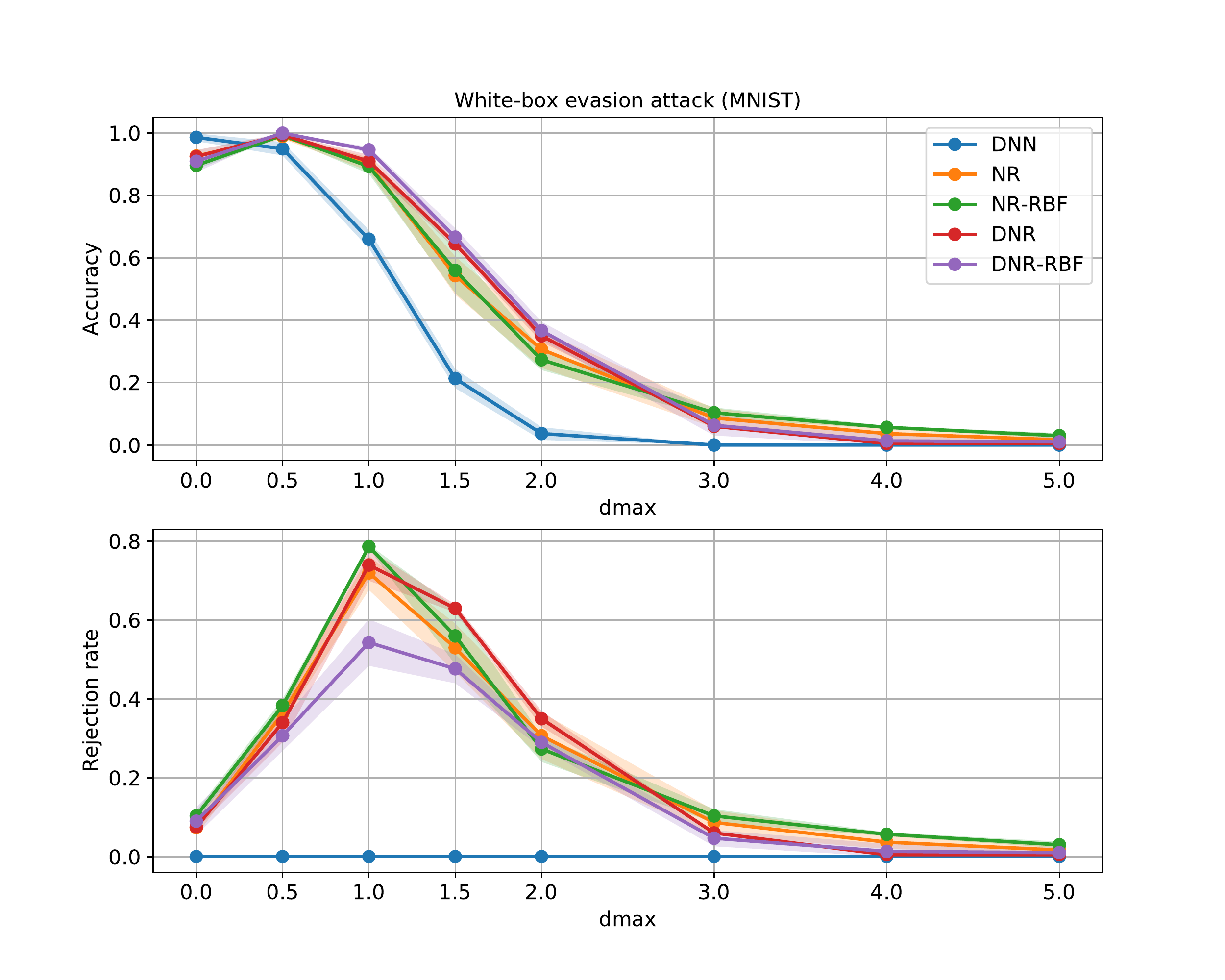}}
    \fbox{\includegraphics[width=.45\textwidth, trim=40 10 50 10, clip]{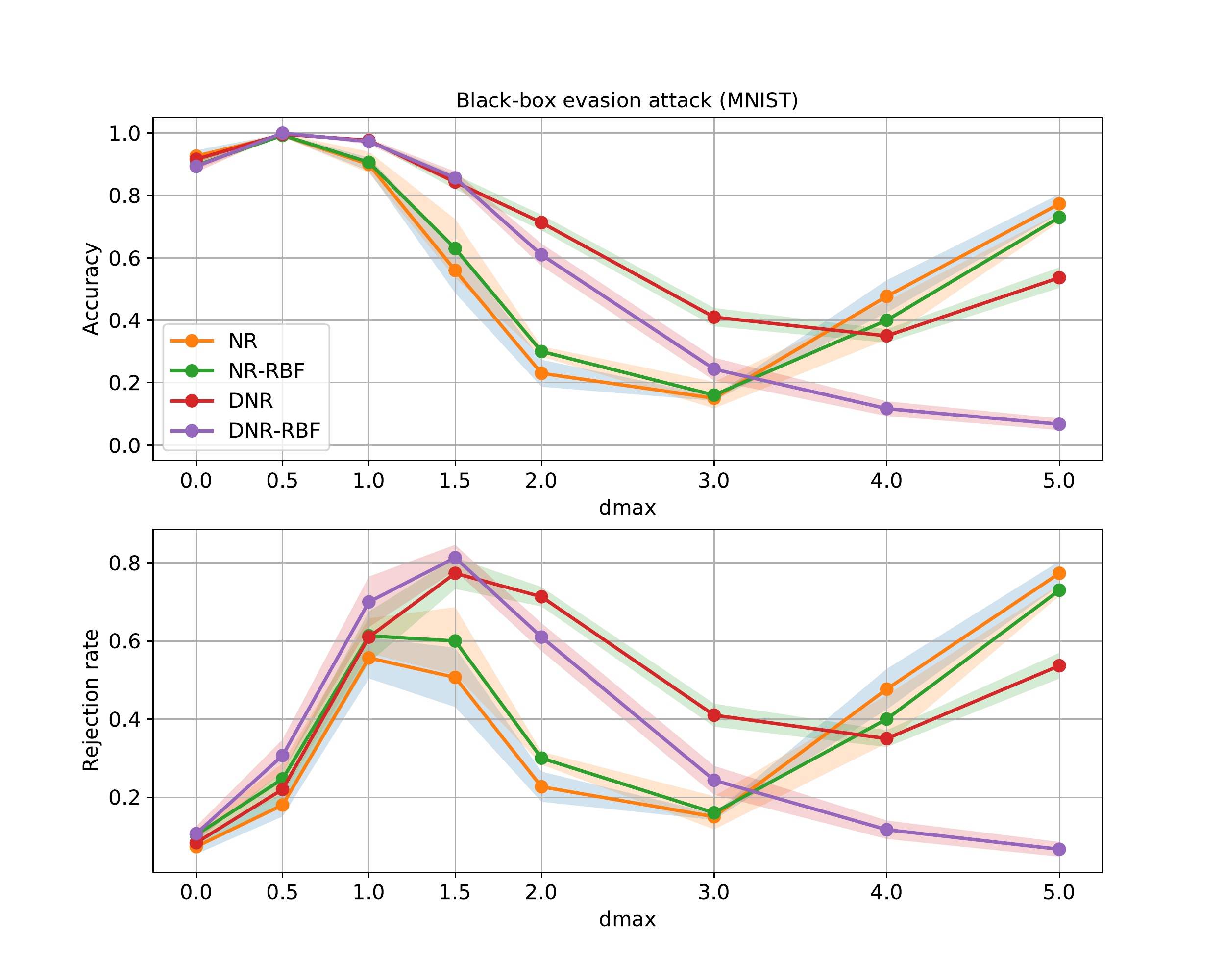}}
    \caption{Security evaluation curves for MNIST data, under white-box (top) and black-box (bottom) settings. Mean accuracy at increasing $\ell_2$-norm perturbation size is reported in the top subplots, while the bottom subplots show the corresponding rejection rates.}
    \label{fig:mnist}
\end{figure}

\begin{figure}[t!]
    \centering
    \fbox{\includegraphics[width=.45\textwidth, trim=40 10 50 10, clip]{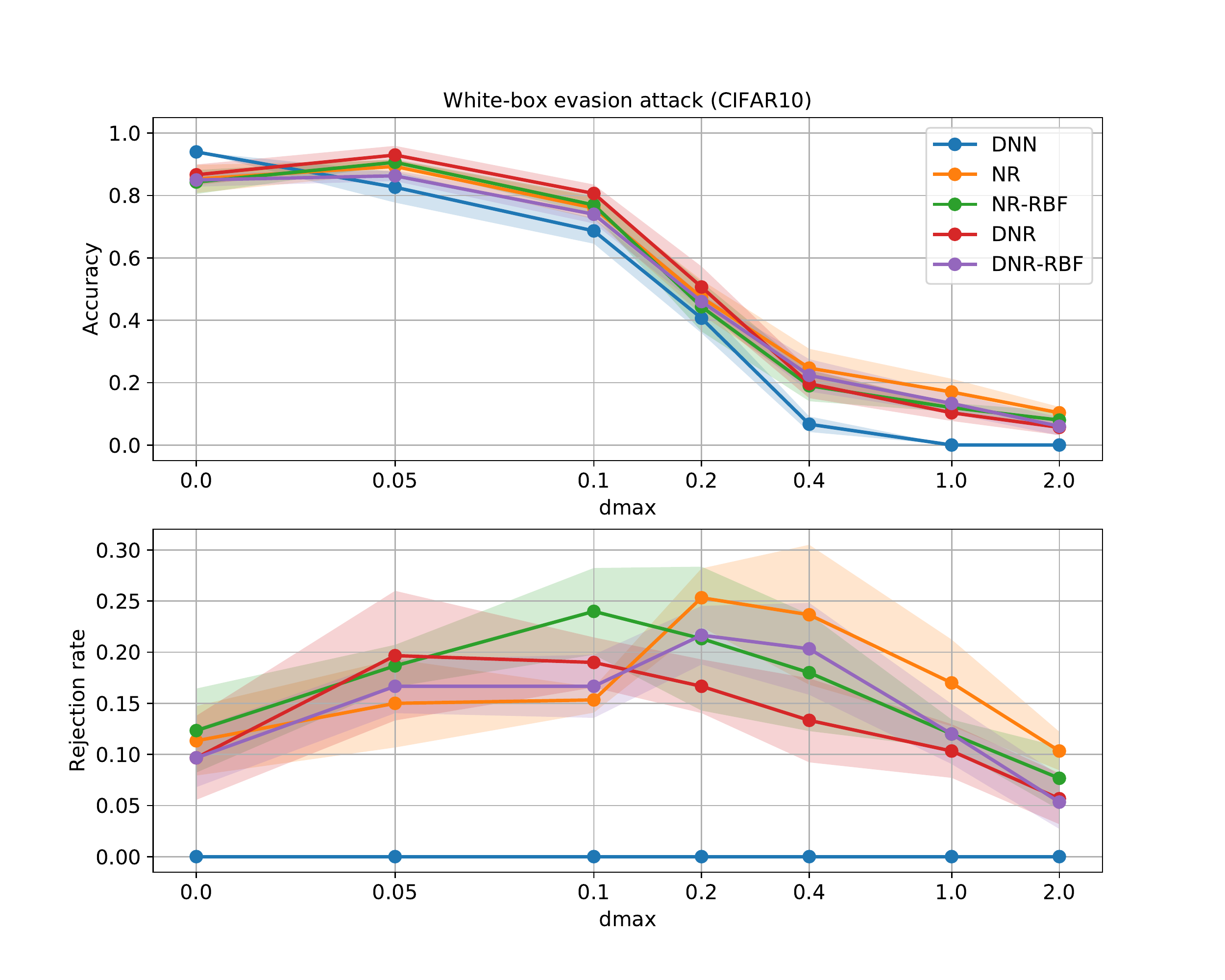}}
    \fbox{\includegraphics[width=.45\textwidth, trim=40 10 50 10, clip]{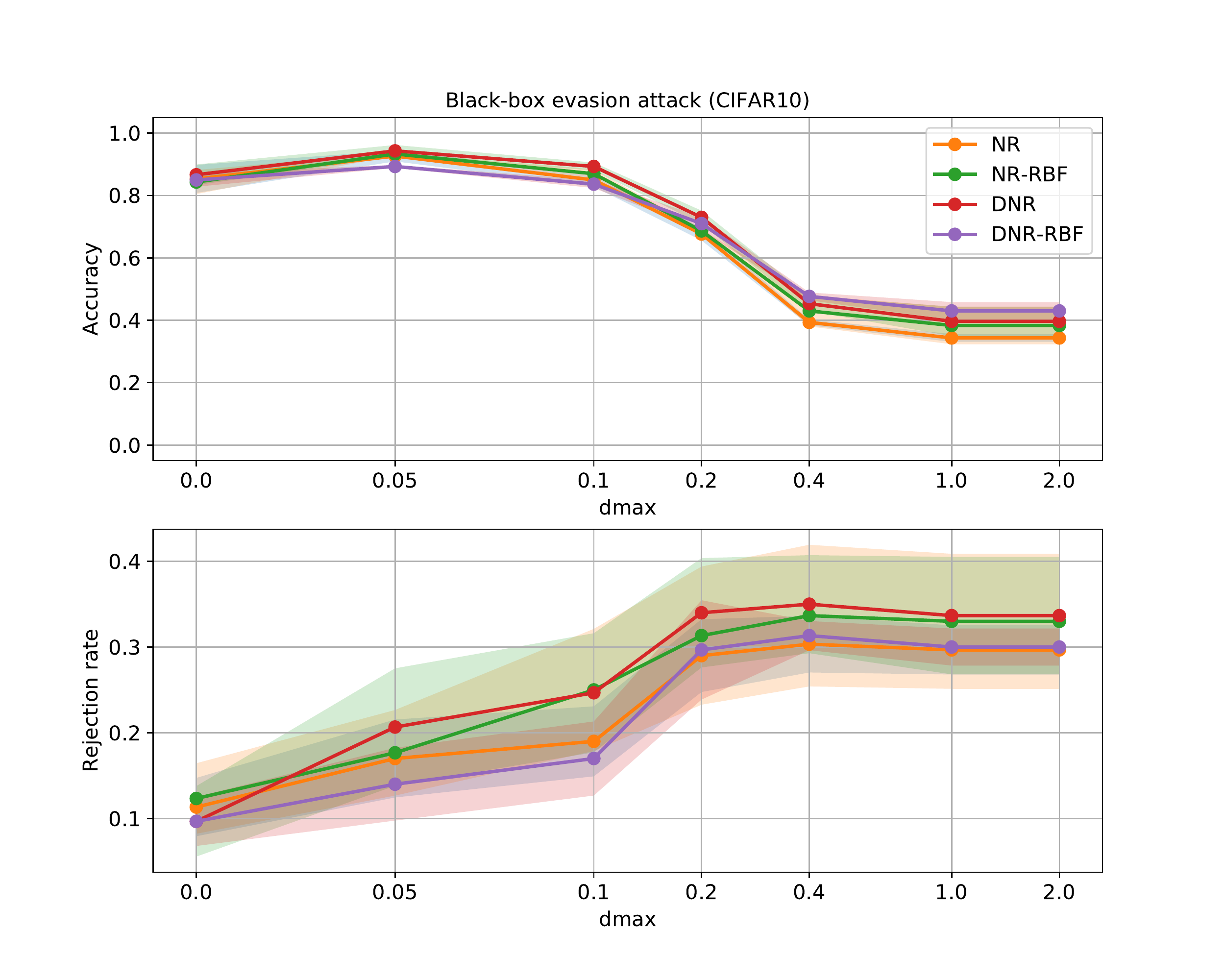}}
    \caption{Security evaluation curves for CIFAR10 data, under white-box (top) and black-box (bottom) settings. Mean accuracy at increasing $\ell_2$-norm perturbation size is reported in the top subplots, while the bottom subplots show the corresponding rejection rates.}
    \label{fig:cifar10}
\end{figure}

The security evaluation curves for both white-box (defense-aware) and black-box settings, against MNIST and CIFAR10 datasets, are reported in Figures \ref{fig:mnist} and \ref{fig:cifar10}, top subplots. In the bottom subplots, we report the corresponding rejection rates at increasing $\ell_2$-norm perturbation size.

In absence of an attack ($\varepsilon = 0$), the undefended DNN slightly outperforms all rejection-based detectors, due to a portion of samples that are incorrectly flagged as positives. Under attack, ($\varepsilon > 0$) all detectors show improved robustness to adversarial examples compared to the standard DNN, as their accuracy decreases more gracefully. Notably, the performance of all detectors even increases for low values of $\varepsilon$, as the slightly modified testing images immediately become blind-spot adversarial examples, ending up in a region which is far from the rest of the data. As the input perturbation increases, such samples are gradually drifted inside a different class, making them indistinguishable for the rejection-based defense \cite{melis17-vipar}.
Interestingly, for the MNIST case, we notice a similar increase of accuracy for NR, DNR, and NR-RBF detectors tested in black-box setting at the highest values of $\varepsilon$. Due to the limited knowledge of the attacker, at high $\ell_2$ distances the adversarial samples end up again once farther from the rest of the data and are rejected by the defenses. This behavior is also confirmed by the corresponding rejection rate, which increases jointly with $\varepsilon$. 

By comparing the average adversarial robustness of the different defense architectures, we show that FADER variants successfully provide a detection accuracy comparable to original DNR and NR counterparts, while dramatically reducing the computational complexity at runtime. Tables \ref{tab:mnist-proto} and \ref{tab:cifar10-proto}, report for MNIST and CIFAR10 datasets respectively, the number of prototypes employed in each detector component along with the estimated prototypes reduction rate for FADER-based defenses.

Notably, up to $73\times$ prototypes reduction rate can be achieved for NR and $20\times$ for DNR without performance drops on both natural and adversarial data for MNIST dataset. For CIFAR10 dataset, \emph{faster} NR (up to $50\times$ prototypes reduction rate) and DNR ($28\times$ prototype reduction rate) are obtained employing FADER with no performance drops on both clean and adversarial data.

\begin{table}[!h]
	\centering
    \resizebox{\columnwidth}{!}{%
	\begin{tabular}{ cccccccc }
		\toprule
		Detector & \multicolumn{6}{c}{\# prototypes}  & Accuracy \\ \cmidrule{2-7}
		& \texttt{relu2} & \texttt{relu3} & \texttt{relu4} & \texttt{combiner} & \texttt{total} & \texttt{reduction} \\
		\midrule
		NR & - & - & 736 & - & 736 & - & 0.984 \\
		\textbf{NR-RBF} & - & - & \textbf{10} & - & \textbf{10} & $\mathbf{\sim73\times}$ & \textbf{0.985} \\
		DNR & 2304 & 2375 & 736 & 9152 & 11527 & - & 0.961 \\
		\textbf{DNR-RBF} & \textbf{250} & \textbf{250} & \textbf{50} & \textbf{10} & \textbf{560} & $\mathbf{\sim20\times}$ & \textbf{0.989} \\
		\bottomrule
	\end{tabular}
	}
	\caption{Comparison of the number of prototypes used by each component of the rejection-based defense architectures (FADER in bold) on MNIST dataset. We also report the mean accuracy of each detector at $\varepsilon = 0$.}
	\vspace{-1em}
    \label{tab:mnist-proto}
\end{table}

\begin{table}[!ht]
	\centering
    \resizebox{\columnwidth}{!}{%
	\begin{tabular}{ cccccccc }
		\toprule
		Detector & \multicolumn{6}{c}{\# prototypes}  & Accuracy \\ \cmidrule{2-7}
		& \texttt{relu2} & \texttt{relu3} & \texttt{relu4} & \texttt{combiner} & \texttt{total} & \texttt{reduction} \\
		\midrule
		NR & - & - & 5257 & - & 5275 & - & 0.915 \\
		\textbf{NR-RBF} & - & - & \textbf{100} & - & \textbf{100} & $\mathbf{\sim50\times}$ & \textbf{0.911} \\
		DNR & 7198 & 3100 & 5257 & 10000 & 25555 & - & 0.913 \\
		\textbf{DNR-RBF} & \textbf{500} & \textbf{300} & \textbf{100} & \textbf{100} & \textbf{900} & $\mathbf{\sim28\times}$ & \textbf{0.892} \\
		\bottomrule
	\end{tabular}
	}
	\caption{Comparison of the number of prototypes used by each component of the rejection-based defense architectures (FADER in bold) on CIFAR10 dataset. We also report the mean accuracy of each detector at $\varepsilon = 0$.}
	\vspace{-1em}
    \label{tab:cifar10-proto}
\end{table}

\section{Related Work}\label{sec:related}
The problem of reactively or proactively countering adversarial attacks is far from being new. The first adversary-aware classification algorithm against evasion attacks has been proposed in 2004, which is based on simulating attacks and iteratively retraining the classifier on them \cite{dalvi04-kdd}. More recently, similar techniques took the name of \emph{adversarial training} and were employed to counter adversarial examples in DNNs \cite{szegedy14-iclr,goodfellow15-iclr}, or to harden decision trees and random forests \cite{kantchelian16-icml}.

As retraining-based techniques are founded on heuristics, with no formal guarantees on convergence and robustness properties, more structured approaches rely on game theory. Zero-sum games learn invariant transformations like feature insertion, deletion and rescaling \cite{globerson06-icml,teo08,dekel10}. Also, other works introduced Nash and Stackelberg games for secure learning, deriving formal conditions for existence and uniqueness of the game equilibrium, under the assumption that each player knows everything about the opponents and the game \cite{bruckner12,liu10a}, or by randomizing players \cite{bulo17-tnnls} and uncertainty on the players’ strategies \cite{grosshans13}. Regrettably, however, machine learning in adversarial scenarios is not a board game with well-defined rules, thus understanding the extent to which the resulting attack strategies are representative of practical scenarios remains an open issue \cite{wooldridge12,cybenko12}. Also, the scalability of these methods to large datasets and high-dimensional feature spaces is in doubt, as it may be too computationally costly to generate a sufficient number of attack samples to correctly represent their distribution.

Another line of research on adversarial defenses takes the name of \emph{robust optimization}. It formulates machine learning in adversarial settings as a min-max problem in which the inner problem maximizes the training loss by manipulating the training points under worst-case, bounded perturbations, while the outer problem minimizes the corresponding worst-case training loss \cite{goodfellow15-iclr,madry2018towards}. A direct result derived from these techniques is the equivalence between regularized learning problems and robust optimization, which has enabled approximating computationally demanding secure learning models, like the aforementioned ones based on game theory, with more efficient strategies based on regularizing the objective function in a specific manner \cite{demontis19-tdsc}. The main effect of these methods is to smooth out the decision function of the classifier reducing the norm of the input gradients, making it less sensitive to worst-case input changes. To achieve this, few works proposed to improve the so-called \emph{evenness} of the classifier's parameters \cite{kolcz09,biggio10-ijmlc,melis2020gradient}.

\section{Conclusions and Future Work}\label{sec:conclusions}
In this work, we presented FADER (\emph{Fast Adversarial Example Rejection}), a technique to speedup rejection-based defenses against adversarial examples. FADER exploits RBF networks to control the number of reference prototypes required for predictions, resulting in accuracy vs. detection time efficiency gain. In our experiments, we demonstrated a $73\times$ prototypes reduction with respect to analyzed detectors for MNIST dataset, and up to an $80\times$ prototypes reduction for CIFAR10 image recognition task, while maintaining comparable performance on both clean and adversarial data. This can have a strong impact on real-world scenarios involving adversarial examples detection on low capability (e.g., edge) devices.

We further provided a comprehensive review of multiple detector-based adversarial detection techniques from the literature, framing them in the form of a proposed adversarial examples detection framework designed to accommodate both existing and newer methods to come (Section \ref{sec:framework}). To demonstrate this, we could frame FADER as well as existing detector methods in literature in terms of our framework easily.

To properly evaluate FADER's response to adversarial attacks, we designed a novel attack algorithm that takes into account the defense to not overestimate performances under attack (see Section \ref{sec:seval}). Experimental results on different image classification tasks highlight FADER-based defenses as more \emph{efficient} solutions than original ones in terms of required prototypes, while maintaining comparable performances both on clean data and under attack.

As future work, we aim to improve FADER performance under attack by employing proper input gradient regularization \cite{demontis19-tdsc, Simon19_iclr} and to test novel FADER architectural variants to further reduce detectors computational overhead.

\section*{Acknowledgements}
This work has been partially supported by the PRIN 2017 project RexLearn (grant no. 2017TWNMH2), funded by the Italian Ministry of Education, University and Research; and by BMK, BMDW, and the Province of Upper Austria in the frame of the COMET Programme managed by FFG in the COMET Module S3AI.

\bibliography{bibDB, FC}

\end{document}